\documentclass[letterpaper, 10 pt, conference]{ieeeconf}  

\IEEEoverridecommandlockouts                              

\usepackage{graphics} 
\usepackage{epsfig} 
\usepackage{mathptmx} 
\usepackage{times} 
\usepackage{amsmath} 
\usepackage{amssymb}  
\usepackage{cite}
\usepackage{graphicx}
\usepackage{tikz}
\usepackage{float}
\usepackage{url}
\usepackage{subcaption}
\usepackage{booktabs}
\usepackage{balance}
\makeatletter\let\labelindent\relax\makeatother
\usepackage{enumitem}
\setlist{nosep, leftmargin=*, topsep=0pt}
\makeatletter
\let\NAT@parse\undefined
\makeatother
\usepackage{hyperref}
\hypersetup{
    colorlinks=true,
    linkcolor=black,     
    filecolor=magenta,
    urlcolor=black,
    citecolor=black,     
    pdftitle={More Is Not Always Better: Sensor Modalities in ACT},
    pdfpagemode=FullScreen,
}
\title{\LARGE \bf
Active Stereo-Camera Outperforms Multi-Sensor Setup in ACT Imitation Learning for Humanoid Manipulation
}
\newif\ifdoubleblind
\doubleblindfalse 

\ifdoubleblind
    \author{Anonymous Author(s)\\
    \textit{Paper under Double-Blind Review}}
\else
    \author{Robin Kühn$^{1}$, Moritz Schappler$^{1}$, Thomas Seel$^{1}$ and  Dennis Bank$^{1}$
    \thanks{$^{1}$The authors are with the Institute of Mechatronic Systems, Leibniz University Hannover, 30167 Hannover, Germany.
            {\tt\small robin.kuehn@imes.uni-hannover.de},
            {\tt\small \{dennis.bank, moritz.schappler, thomas.seel\}@imes.uni-hannover.de}}%
    }
\fi

\begin{document}

\maketitle
\thispagestyle{empty}
\pagestyle{empty}

\bstctlcite{IEEEexample:BSTcontrol}

\begin{abstract}
The complexity of teaching humanoid robots new tasks is one of the major reasons hindering their widespread adoption in the industry. While Imitation Learning (IL), particularly Action Chunking with Transformers (ACT), enables rapid task acquisition, there is no consensus yet on the optimal sensory hardware required for manipulation tasks. 
This paper benchmarks 14 sensor combinations on the Unitree G1 humanoid robot equipped with three-finger hands for two manipulation tasks. We explicitly evaluate the integration of tactile and proprioceptive modalities alongside active vision. Our analysis demonstrates that strategic sensor selection can outperform complex configurations in data-limited regimes while reducing computational overhead. We develop an open-source \textit{Unified Ablation Framework} that utilizes sensor masking on a comprehensive master dataset.
Results indicate that additional modalities often degrade performance for IL with limited data. A minimal active stereo-camera setup outperformed complex multi-sensor configurations, achieving 87.5\% success in a spatial generalization task and 94.4\% in a structured manipulation task. Conversely, adding pressure sensors to this setup reduced success to 67.3\% in the latter task due to a low signal-to-noise ratio.
We conclude that in data-limited regimes, active vision offers a superior trade-off between robustness and complexity. While tactile modalities may require larger datasets to be effective, our findings validate that strategic sensor selection is critical for designing an efficient learning process.
\ifdoubleblind
\else
    Code is available at~\url{http://github.com/kuehnrobin/UAF_unitree_g1}.
\fi
\end{abstract}

\section{Introduction}

While classical robots lack the flexibility to operate in unstructured environments, humanoids operate in human-centric spaces without requiring infrastructure modifications~\cite{shamsuddohaHumanoidRobotsTesla2025}. However, programming complex manipulation remains challenging. 
Imitation Learning (IL) approaches like Action Chunking with Transformers (ACT)~\cite{zhao2023learningfinegrainedbimanualmanipulation} address this hurdle by enabling robots to learn fine-grained manipulation from demonstrations~\cite{buamanee2024biactbilateralcontrolbasedimitation}.

Despite the success of recent IL frameworks~\cite{fu2024mobile, fu2024humanplus}, there is no consensus on the optimal sensory hardware required. While works like Bi-ACT~\cite{buamanee2024biactbilateralcontrolbasedimitation, kobayashiALPHABiACTAre2025} suggest the utility of joint torques, and OpenTelevision~\cite{cheng2024tv} advocates for active vision, these modalities are typically validated without rigorous comparison to alternative setups. Recent frameworks like ULTRA~\cite{heULTRAUnifiedMultimodal2026} highlight the ongoing, complex efforts to unify multimodal sensor streams for autonomous behavior, underscoring the need for a foundational understanding of which sensors actually benefit the learning process.

Crucially, these modalities are evaluated across diverse robotic platforms with varying sensor fidelity. For instance, Mobile ALOHA utilizes ViperX/WidowX arms~\cite{fu2024mobile}, OpenTelevision evaluates on full-size humanoids (Unitree H1, Fourier GR-1)~\cite{cheng2024tv}, and Bi-ACT relies on the Dynamixel-based ALPHA-${\alpha}$ platform~\cite{kobayashiALPHABiACTAre2025}. Consequently, it remains unclear whether reported performance gains stem from the inherent value of the added modalities or merely from specific hardware characteristics. Classically, evaluating different sensor setups requires recording separate datasets or comparing results across different research papers. Both approaches introduce varying operator skill and demonstration quality as hidden variables, which diminishes the validity of the comparison. 

To bridge this gap, a unified benchmark on a single platform is required. While an evaluation on a single platform cannot claim universal transferability to high-fidelity, custom-built robots, it is uniquely suited for a controlled ablation study to isolate the impact of sensor selection. We purposefully conduct this evaluation on the Unitree G1 (version Edu-U4) equipped with standard three-finger hands, as it is representative of the emerging class of cost-effective, commercially available humanoids. As the community increasingly adopts such platforms, integrating their inherently noisy, lower-fidelity sensors presents unique challenges for imitation learning. Identifying the minimal viable sensor set here not only improves robustness but also reduces required computational resources and hardware costs.

To achieve this, we benchmark 14 distinct sensor configurations across two complementary manipulation tasks. Unlike prior studies, we explicitly investigate the integration of built-in finger-pressure sensors and torque-based proprioception alongside active and passive vision. To ensure strict comparability across these setups, we introduce an open-source \textit{Unified Ablation Framework} (UAF). This pipeline trains multiple policy variations from a single master dataset via sensor masking, effectively eliminating human demonstration variance as a confounding factor. By training all policies on identical human trajectories, our framework guarantees that performance differences are uniquely attributable to the chosen sensory input.


The main contributions of this work are:
\begin{enumerate}
    \item \textbf{Unified Ablation Framework (UAF):} We introduce an open-source toolchain utilizing sensor masking on a shared master dataset, building upon the LeRobot Framework~\cite{cadene2024lerobot}. This allows for a comparison of sensor configurations by guaranteeing identical training episodes regarding the human demonstrations. 
    \ifdoubleblind
\else
        The UAF is made publicly available at~\url{https://github.com/kuehnrobin/UAF_lerobot}.
    \fi
    \item \textbf{Extensive Evaluation of Sensor Modalities:} We provide an extensive evaluation of the impact of different sensory combinations on the performance of ACT IL. This includes proprioception by utilizing the pressure sensors from the Unitree Dex3 hands~\cite{unitree_dex3_1_2025}. The collected training data for both evaluated tasks is made open-source and represents typical \textit{data-limited regimes} (defined here as~$\le 250$ demonstrations).
    \ifdoubleblind
\else
        Specifically, the \textit{sort cans} task consists of 20 episodes (approx. 1 hour of demonstration data)~\cite{kuehn2026cans}, and the \textit{place cubes in box} task consists of 242 episodes (approx. 2 hours of demonstration data)~\cite{kuehn2026cubes}.
    \fi
    \item \textbf{Minimalist Design Guidelines:} We validate that a single active stereo-camera is often sufficient for designing an efficient learning process for ACT-based IL for humanoid manipulation in such data-limited regimes. 
\end{enumerate}

After introducing related work (Sec.~\ref{sec:related_work}), methodology (Sec.~\ref{sec:methodology}) and discussing results (Sec.~\ref{sec:results}), the latter point of practical design guidelines is elaborated in Sec.~\ref{sec:guidelines}. This is followed by a discussion of limitations in Sec.~\ref{sec:limitations} and concluded in Sec.~\ref{sec:conclusion}.

\section{RELATED WORK}
\label{sec:related_work}

\subsection{Imitation Learning for Humanoid Manipulation}
Imitation Learning (IL) has evolved from simple Behavior Cloning to sophisticated sequence modeling. Action Chunking with Transformers (ACT)~\cite{zhao2023learningfinegrainedbimanualmanipulation} and Diffusion Policies~\cite{chiDiffusionPolicyVisuomotor2024, zeGeneralizableHumanoidManipulation2025} have emerged as dominant architectures. Subsequent frameworks have scaled these methods to complex whole-body humanoid control~\cite{fu2024humanplus} and bimanual mobile manipulation~\cite{fu2024mobile}. To facilitate data collection for these data-hungry models, recent works explore robot-free demonstration interfaces~\cite{naiHumanoidManipulationInterface2026} or implicit kinodynamic motion retargeting~\cite{chenImplicitKinodynamicMotion2025a}. However, while frameworks like OpenTelevision~\cite{cheng2024tv} compare different visual sensory modalities, the question of the minimal sufficient sensor set for stationary manipulation tasks on standard hardware remains unanswered.

\subsection{Active Perception in Imitation Learning}
Active vision has gained traction in IL through frameworks like OpenTelevision~\cite{cheng2024tv} and Vision-in-Action~\cite{xiong2025visionactionlearningactive}. Unlike classical active perception, which often relies on heuristics, these approaches learn active perceptual strategies end-to-end from human demonstrations. The efficacy of pure vision-based pipelines on the Unitree G1 has also been demonstrated in highly dynamic tasks, such as perceptive parkour~\cite{wuPerceptiveHumanoidParkour2026a}. This decoupling of the camera view from the torso notably improves performance compared to static baselines. However, current research validates these learned active perception policies primarily in isolation. Our work extends this by exploring interference effects when combining active vision with static wide-angle views.

\subsection{Tactile and Proprioceptive Feedback in IL}
While visual inputs are standard in Transformer-based IL, the integration of haptics and proprioception has only recently gained traction. Methods like Bi-ACT~\cite{buamanee2024biactbilateralcontrolbasedimitation} utilize joint velocities and estimated forces for force-sensitive tasks. To capture high-fidelity contact dynamics, recent systems propose specialized hardware. Choi et al. employ a custom 6-DoF force/torque sensor at the fingertips in UMI-FT~\cite{choiIntheWildCompliantManipulation2026a}, while TACT~\cite{murookaTACTHumanoidWholebody2025} extends ACT with advanced tactile sensor arrays across the robot's body to stabilize whole-body manipulation. However, there is a noteworthy gap between these custom, high-end sensor suites and the standard hardware of emerging commercial humanoids, which typically rely on basic fingertip pressure sensors and direct motor torques. The utility of integrating these standard, built-in modalities into ACT policies remains largely unexplored. We address this gap by evaluating whether the inclusion of these accessible tactile and proprioceptive sensors provides a tangible performance benefit for manipulation tasks in data-limited regimes.

\section{METHODOLOGY}
\label{sec:methodology}

To isolate the influence of sensor selection on policy performance, we developed a standardized experimental framework using the Unitree G1 humanoid robot~\cite{unitree_g1, unitree_g1_docs}.

\subsection{Scope of the Data Regime}
Before detailing the system architecture, it is crucial to define the scope of our evaluation. Recent advancements in robotic manipulation span a wide spectrum of data regimes. While foundation models leverage massive datasets exceeding 100,000 episodes to learn generalized representations~\cite{brohanRT2VisionLanguageActionModels2023}, single-task Imitation Learning typically relies on 50 to 250 demonstrations~\cite{zhao2023learningfinegrainedbimanualmanipulation, fu2024mobile, chiDiffusionPolicyVisuomotor2024}. 

In this work, we explicitly target this \textit{data-limited regime} (defined as~$\le 250$ demonstrations), as it represents the most realistic scenario for rapid, on-the-fly skill acquisition without massive data collection infrastructure or the need for large-scale robot-free approximations~\cite{naiHumanoidManipulationInterface2026}.

\subsection{Augmented ACT Architecture}
We utilize the Unitree wrapper~\cite{unitree_il_lerobot} of the \textit{LeRobot} library~\cite{cadene2024lerobot} as the foundational codebase for our implementation of ACT~\cite{zhao2023learningfinegrainedbimanualmanipulation}. The model operates as a Conditional Variational Autoencoder (CVAE) predicting action sequences~$a_{t:t+k}$ based on current observations and a latent style variable~$z$. To ensure comparability with state-of-the-art baselines, we adopted the hyperparameter set from OpenTelevision~\cite{cheng2024tv} and utilized a ResNet18 visual backbone for all policies.

\begin{figure}[t]
  \centering
  \includegraphics[width=\columnwidth]{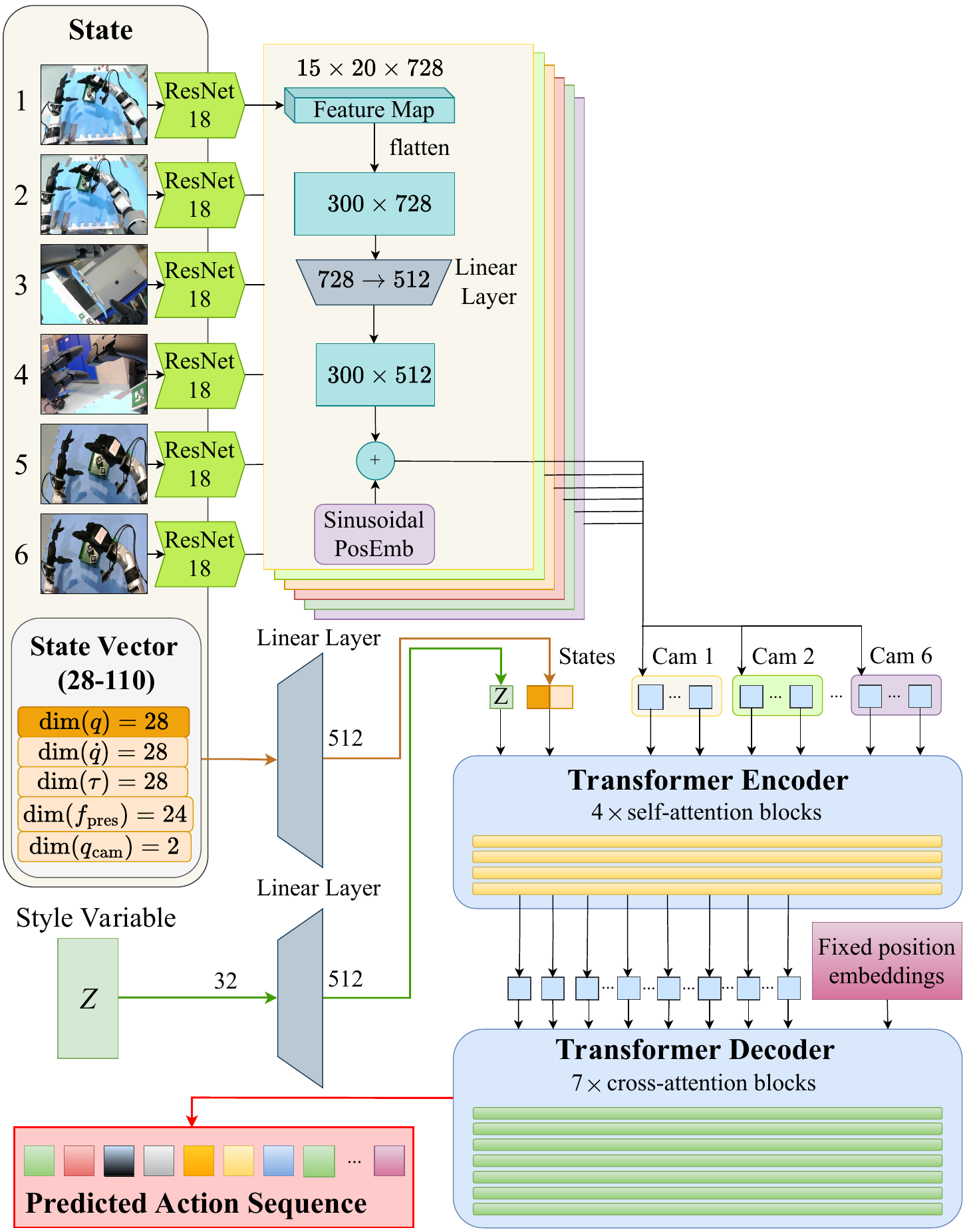}
    \caption{\textbf{Custom ACT Architecture:} Schematic representation integrating pressure sensors, joint velocities, and torques into the state observation vector. Building upon:~\cite{zhao2023learningfinegrainedbimanualmanipulation, cadene2024lerobot, cheng2024tv}.}
  \label{fig:act_arch}
\end{figure}

\textbf{State-Space Augmentation:} 
We extended the standard LeRobot ACT implementation to obtain a configurable state vector (Fig.~\ref{fig:act_arch}). Unlike the default implementation, which typically relies solely on joint positions, our modified proprioceptive encoder accepts an augmented vector~$s_t$. This vector concatenates joint positions~$q \in \mathbb{R}^{n}$, velocities~$\dot{q} \in \mathbb{R}^{n}$, motor torques~$\tau \in \mathbb{R}^{n}$, and tactile pressure data~$f_{\text{pres}} \in \mathbb{R}^{m}$ from the fingertips:
\begin{equation}
    s_t = [q, \dot{q}, \tau, f_{\text{pres}}]
\end{equation}
These inputs are concatenated and projected via a learnable linear layer to match the transformer's embedding dimension.

\subsection{Teleoperation and Active Perception}
We implemented a VR-based teleoperation system building upon OpenTelevision and Unitree's AVP\_Teleoperate~\cite{cheng2024tv, unitreerobotics_avp_teleoperate}, using the Meta Quest 3 headset~\cite{metaQuest3}. While recent systems like HumDex~\cite{hengHumDexHumanoidDexterous2026} expand the G1's capabilities by integrating highly complex 20-DoF dexterous hands, we deliberately utilize the stock \textbf{7-DoF Dex3-1 hands}~\cite{unitree_dex3_1_2025} to benchmark the baseline sensory requirements of commercially available humanoids. 

\begin{figure}[t]
  \centering
  \includegraphics[width=\columnwidth]{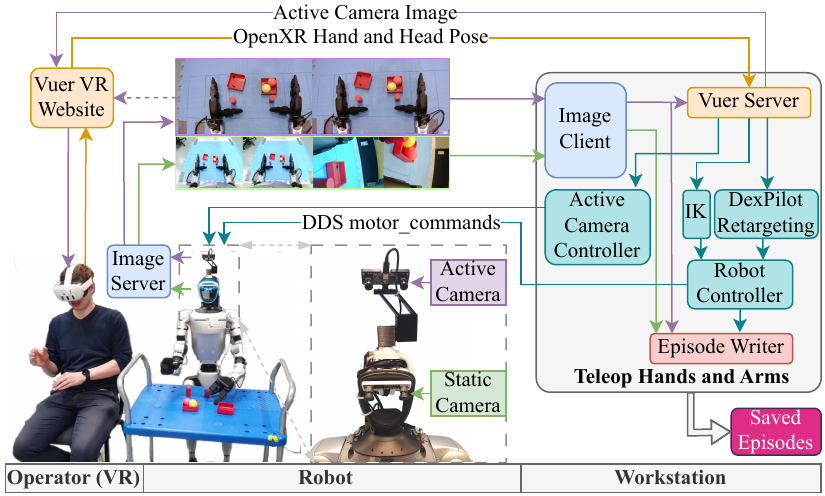}
  \caption{\textbf{Teleoperation Setup:} The operator controls the Unitree G1 (with three-finger hands) via VR. Head movements are synchronized, enabling the collection of active vision data.}
  \label{fig:teleoperation_setup}
\end{figure}

Our system retargets human hand motions using a modified DexPilot algorithm~\cite{handa2019dexpilotvisionbasedteleoperation, qin2023handmultiplehandsimitation}. Crucially, the operator views the environment directly through the robot's active stereo-camera (Fig.~\ref{fig:teleoperation_setup}). This creates a causal link between head movement and visual observation, enabling the ACT policy to learn \textit{active perception} strategies (e.g., visual search) from demonstrations.

\subsection{Unified Ablation Framework}
\label{sec:unified_ablation_framework}

\begin{figure}[t]
  \centering
  \includegraphics[width=\columnwidth]{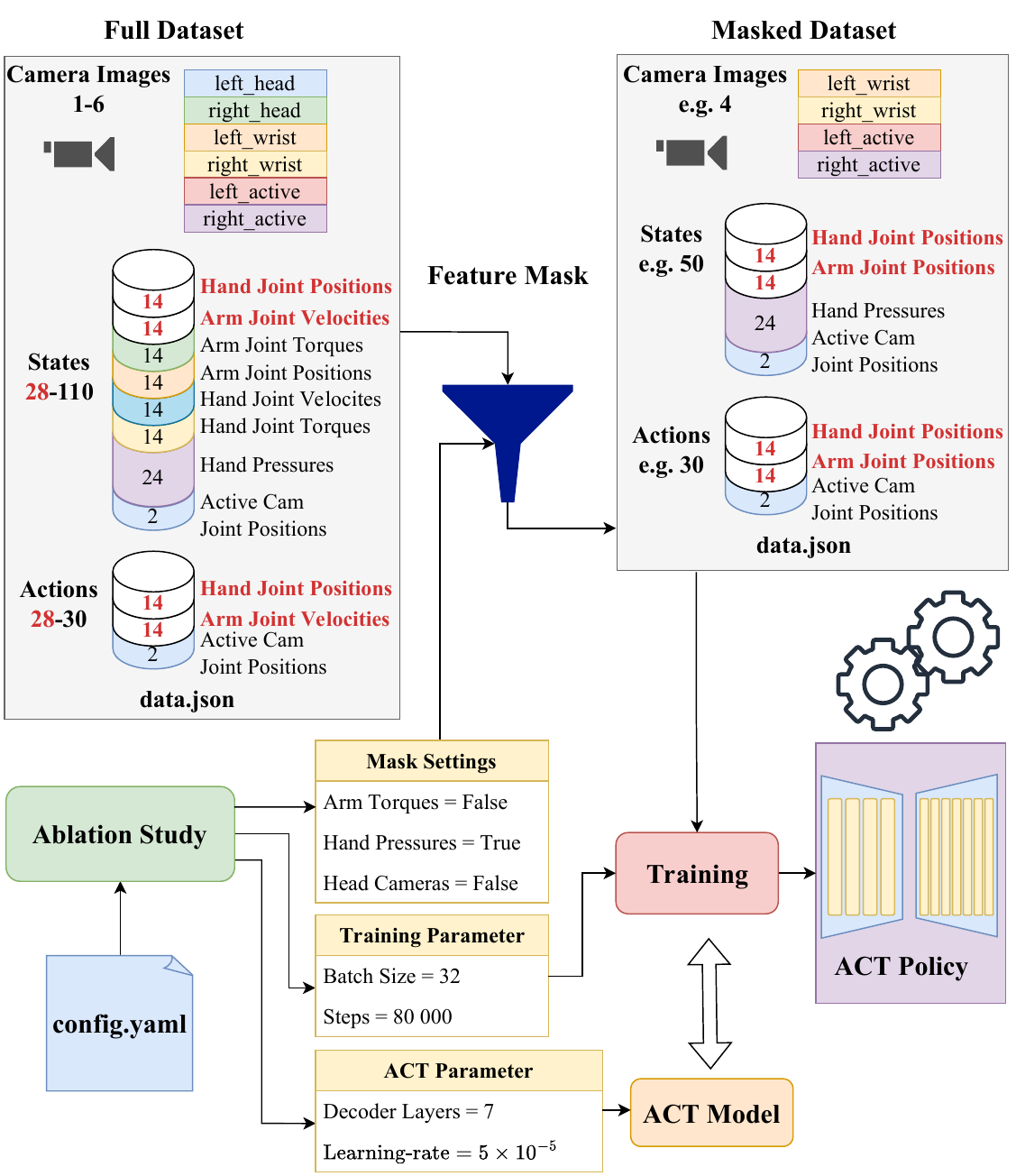}
    \caption{\textbf{Unified Ablation Framework:} To prevent human variance from influencing results, we record a master dataset with all sensors. During training, a masking module creates specific configurations (e.g.,~$A$,~$WA-P$) from identical demonstration data.}
  \label{fig:ablation_pipeline}
\end{figure}

To systematically evaluate the impact of sensor selection, we integrated a \textit{Unified Ablation Framework} directly into the LeRobot training pipeline (Fig.~\ref{fig:ablation_pipeline}).

\textbf{Master Dataset \& Sensor Masking:} 
All demonstrations are recorded into a comprehensive dataset encompassing every available modality (Active Cam, Static Cam, Wrist Cams,~$q, \dot{q}, \tau, f_{\text{pres}}$). To emulate specific hardware setups, a custom data loader applies a binary mask to the observation vector during training, dynamically omitting excluded sensor streams. This strictly enforces perfectly synchronized training episodes across all evaluated policies.

\textbf{Policy Naming Convention:} 
We employ the following policy-naming scheme: Camera Setup~($\underline{A}$ctive,~$\underline{S}$tatic,~$\underline{W}$rist) and Proprioception~($\underline{P}$ressure,~$\underline{V}$elocity,~$\underline{T}$orque). When only one camera of a stereo-camera is used, it is denoted by an underscore~$\underline{L}eft$ or~$\underline{R}ight$. For example,~$WA-P$ denotes a policy using Wrist and Active cameras plus Pressure sensors.

\subsection{Experimental Tasks}

\begin{figure}[t]
    \centering
    \begin{minipage}[t]{0.48\columnwidth}
        \centering
        \begin{tikzpicture}[baseline=(image.north)]
            \ifdoubleblind
\else
                \node[anchor=north west, inner sep=0] (image) at (0,0) {
                    \href{https://seafile.projekt.uni-hannover.de/f/fc056a5182c34eaa8c84/}{
                        \includegraphics[width=1.0\linewidth]{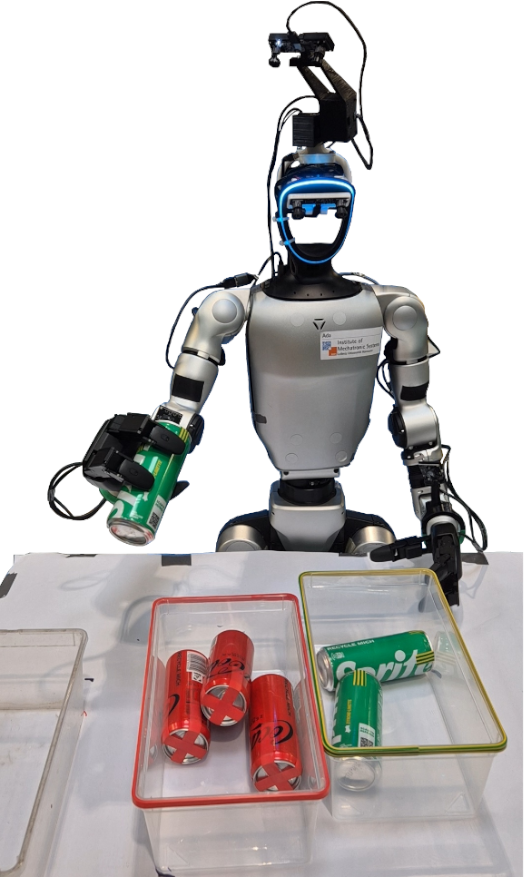}
                    }
                };
                ; 
            \fi
        \end{tikzpicture}
        \caption{Qualitative results of the \textit{Sort Cans} task ($A$ policy).
        \ifdoubleblind
\else
            The picture links to a video of the autonomous task execution.
        \fi
        }
        \label{fig:eval_cans}        
    \end{minipage}
    \hfill 
    \begin{minipage}[t]{0.48\columnwidth}
        \centering
        \begin{tikzpicture}[baseline=(image.north)] 
            \ifdoubleblind
\else
                \node[anchor=north west, inner sep=0] (image) at (0,0) {
                    \href{https://seafile.projekt.uni-hannover.de/f/d40a7d127f5a4941a352/}{
                        \includegraphics[width=1.0\linewidth]{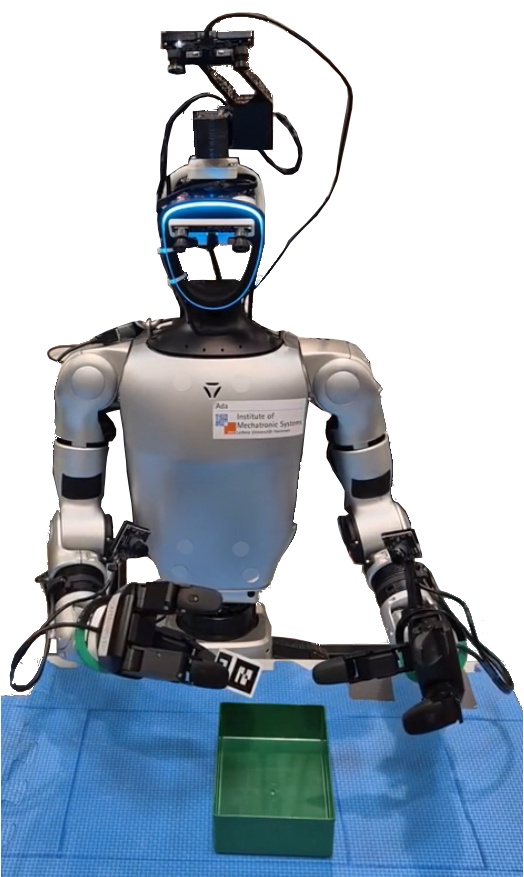}
                    }
                };
                ;
            \fi
        \end{tikzpicture}
        \caption{Qualitative results of the \textit{Grasp Cubes} task ($A$ policy).
        \ifdoubleblind
\else
            The picture links to a video of the autonomous task execution.
        \fi
        }
        \label{fig:eval_cubes}
    \end{minipage}
\end{figure}

We selected two tasks to evaluate different policy attributes (Fig.~\ref{fig:eval_cans},~\ref{fig:eval_cubes}). To account for the variance introduced by randomization, we evaluated each policy over~$N=10$ trials (Task 1, with six picked cans per trial) and~$N=15$ trials (Task 2, with one cube per trial).

\subsubsection{Sort Cans (Baseline Comparison)}
Adopted directly from OpenTelevision~\cite{cheng2024tv}, this task serves as a standardized benchmark. It requires grasping and sorting beverage cans by color at predictable locations.

\subsubsection{Grasp Cubes (Spatial Generalization)}
To test fine manipulation and spatial reasoning beyond the training distribution, we designed a task involving small~($15\times15$mm) cubes (Fig.~\ref{fig:eval_cubes}). Unlike the structured can task, cube positions are randomized across the entire workspace during inference. This specifically challenges the policy's depth perception and active search capabilities.

\subsection{Training Details}
To ensure full reproducibility and facilitate future research, 
\ifdoubleblind
\else
    the corresponding master datasets~\cite{kuehn2026cans, kuehn2026cubes} and the \textit{Unified Ablation Framework} (code at~\url{https://github.com/kuehnrobin/UAF_lerobot}) are made publicly available.
\fi
Model training was conducted on a workstation equipped with a single NVIDIA RTX 4090 (24\,GB) GPU. Utilizing the LeRobot pipeline, each policy was trained with 80,000 optimization steps. The training took 7.3--20.6\,h for the can-sorting ACT runs and 9.6--25.4\,h for the cube-in-box ACT runs, depending on the selected sensor configuration. Since the training schedule in this work is step-based, the equivalent number of epochs varies across sensor configurations (depending on dataset size and batch size). We empirically observed that both the validation loss and the physical behavioral performance stabilize well within this training budget.

\section{RESULTS AND DISCUSSION}
\label{sec:results}

We evaluated 14 distinct sensor configurations across two manipulation tasks to identify the influence of sensor configuration on policy success for small datasets (Table~\ref{tab:results_summary}). The success rate represents the average completion score across five defined subtasks to capture partial progress. The optimal region in the result plots in Fig.~\ref{fig:pareto_combined} is the bottom-right corner, representing low execution time and high success rate.

\begin{table}[t]
    \centering
    \caption{Performance Statistics per Policy and Task. All policies were trained for 80,000 optimization steps. Note: Evaluation counts (Trials) vary because failing policies (e.g., due to hovering) exceeded maximum trial durations, leading to truncated evaluations.}
    \label{tab:results_summary}
    \begin{tabular}{l c c c c}
        \toprule
        \textbf{Policy} & \textbf{Training} & \textbf{Trials} & \textbf{Execution} & \textbf{Success} \\
        & \textbf{Time (h)} & & \textbf{Time (min)} & \textbf{Rate (\%)} \\
        \midrule
        \multicolumn{5}{l}{\textit{\textbf{Task 1: Can Sorting (20 Episodes)}}} \\
        \midrule
        A                           & 7.35 & 11 & \textbf{3.9} & 94.4 \\
        A-P                         & 7.40 & 4  & -    & 67.3 \\
        SW                          & - & 6  & 4.2 & 76.4 \\
        $S_{\text{L}}WA$            & - & 6  & 8.7 & 85.1 \\
        WA                          & 20.58 & 11 & 5.8 & 93.6 \\
        WA-P                        & 19.35& 6  & 4.8 & \textbf{97.6} \\
        $WA\text{-}PV_{\text{A}}T_{\text{A}}$ & 15.35 & 7 & 5.6 & 94.5 \\
        $W_{\text{R}}A$       & 10.32 & 5 & 5.9 & 94.8 \\
        \midrule
        \multicolumn{5}{l}{\textit{\textbf{Task 2: Cube in Box (242 Episodes)}}} \\
        \midrule
        A                           & 11.51 & 15 & \textbf{0.4} & \textbf{87.5} \\
        S                           & 9.56 & 6  & -    & 10.0 \\
        $S_{\text{L}}WA\text{-}P$     & - & 22 & 2.2 & 41.5 \\
        $S_{\text{L}}WA\text{-}PV_{\text{A}}T_{\text{A}}$ & 32.07 & 15 & 2.4 & 56.7 \\
        WA                          & - & 11 & 0.7 & 72.7 \\
        WA-P                        & 13.58 & 20 & 0.7 & 68.1 \\
        $WA\text{-}PV_{\text{A}}T_{\text{A}}$  & 15.35 & 11 & 0.9 & 82.5 \\
        \bottomrule
    \end{tabular}
\end{table}

\subsection{Task 1: Sort Cans (Structured Manipulation)}

\begin{figure}[t]
  \centering
  \includegraphics[width=\columnwidth]{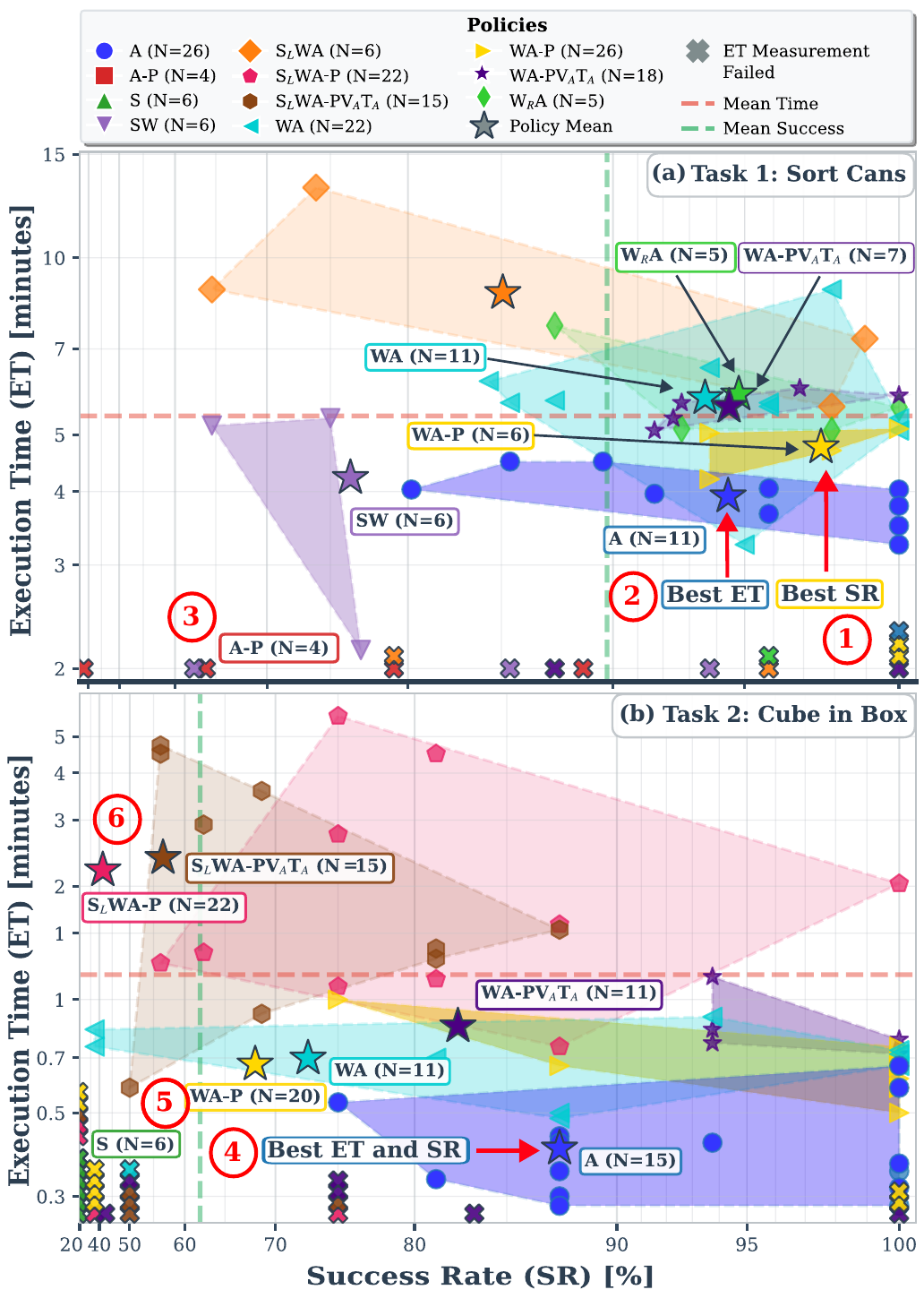}
    \caption{\textbf{Execution time vs. success rate (Task 1):} The maximalist setup~($WA-P$; marked with \textcircled{1} in the plot) achieves the best combination of execution time and success. However, using only a single active stereo-camera~($A$~\textcircled{2}) offers nearly identical results with a fraction of the state-vector and thereby computational complexity. Notably, simply adding pressure sensors~($A-P$~\textcircled{3}) causes a performance drop compared to the camera-only baseline.
    \textbf{Execution time vs. success rate (Task 2):} In generalization tasks, the minimal active setup~($A$~\textcircled{4}) dominates with the combination of the highest success (mean \textbf{87.5\%}) and the fastest execution time with a mean of \textbf{0.4 min}. Conversely, setups involving static cameras (e.g.,~$S$~\textcircled{5},~$S_{\text{L}}WA\text{-}P$~\textcircled{6}) yield suboptimal performance and increased execution times.}
  \label{fig:pareto_combined}
\end{figure}

The \textit{Sort Cans} task represents a structured environment with predictable object locations. Fig.~\ref{fig:pareto_combined}~(a) illustrates the execution time and the success rate of the tested policies.

\textbf{Competitive Performance of Active Vision Setup:}
The \emph{highest absolute performance} was achieved by the complex configuration~$WA-P$~\textcircled{1} (Wrist + Active + Pressure) with a mean \textbf{97.6\%} success rate and a \textbf{4.8 min} mean execution time. This suggests that for structured tasks, the combination of close-up wrist views and tactile feedback can improve robustness.
However, the minimal active stereo-camera policy~($A$~\textcircled{2}) remained \emph{highly competitive} with a mean \textbf{94.4\%} success rate and the fastest mean execution time of \textbf{3.9 min}. 
This minimal setup operates with \emph{a fraction of the state-vector complexity} of multi-sensor configurations, directly translating to lower inference latency and reduced computational cost for training.

\textbf{Comparison to State-of-the-Art:}
To contextualize our performance, we compare our results against the OpenTelevision baseline~\cite{cheng2024tv}, which utilized a similar 'Sort Cans' task. Comparing against their ResNet18 baseline (to ensure backbone parity), they reported an 83\% pick rate and a 50\% place rate. While our policies~$WA-P$~\textcircled{1} and~$A$~\textcircled{2} achieved a higher overall success rate of 97.6\% and 94.4\%, we note that the OpenTelevision baseline was evaluated over only~$N=5$ episodes and did use a slightly different method to determine the success rate~\cite{cheng2024tv}. The main reason for the performance increase is most likely the twofold amount of training data used by us.

\begin{figure}[t]
  \centering
  \includegraphics[width=1.0\columnwidth]{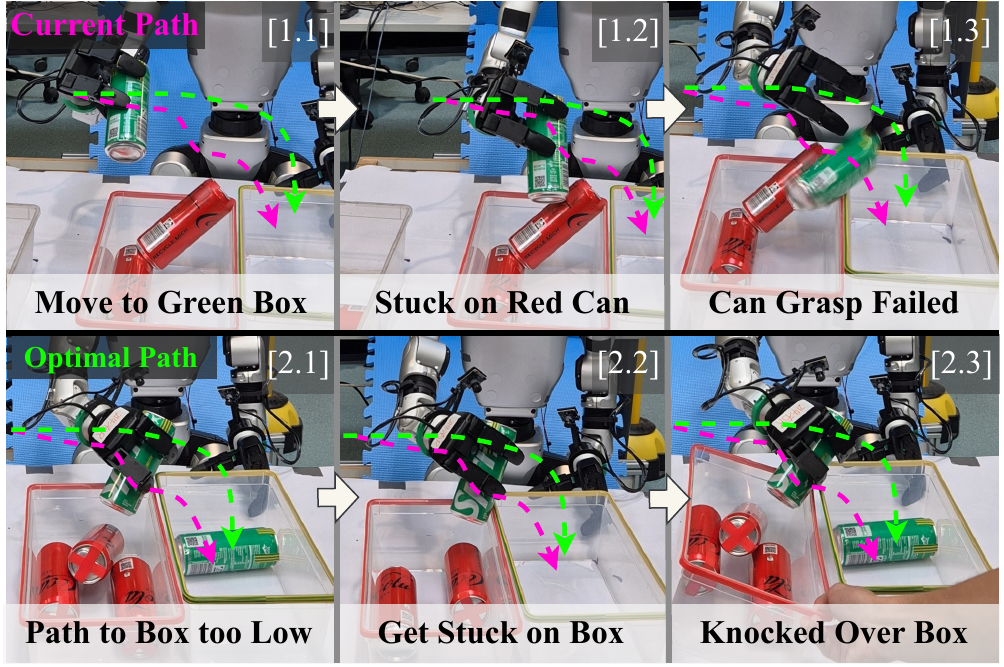}
  \caption{\textbf{Mode Averaging Failure in Task 1:} The policy fails to commit to a single drop-off target (e.g., green vs. red box) and instead generates a path that interpolates between the two distinct modes. This results in a trajectory that is notably lower than the demonstration data, causing the end-effector to collide with the box rim (2.1--2.3) or to collide with previously dropped cans (1.1--1.3).}
  \label{fig:mode_averaging_failure}
\end{figure}

Moreover, we observed a frequent failure mode indicative of \textit{mode averaging}. The policy appeared to interpolate between the distinct trajectories targeting different drop zones, resulting in a path substantially lower than the training distribution and causing frequent collisions with the box rim or previously dropped cans (Fig.~\ref{fig:mode_averaging_failure}). 

\textbf{Low SNR in Tactile Modalities:}
Crucially, adding pressure sensors to the active camera~($A \rightarrow A-P$~\textcircled{3}) caused a \emph{substantial performance drop} to \textbf{67.3\%}. We hypothesize that without sufficient data to densely cover the tactile state-space, inference-time pressure signals deviate from the training distribution due to minor grasp variations. While the hardware sensor itself may possess a high signal fidelity, the policy interprets this out-of-distribution variance as informational noise. This drastically lowers the \emph{effective, task-relevant signal-to-noise ratio} for the learning process, destabilizing the grasp. Tactile data only aided performance when supported by wrist cameras~($WA-P$~\textcircled{1}) to provide sufficient visual context to disambiguate the signals.

\subsection{Task 2: Grasp Cubes (Spatial Generalization)}
\label{sec:cubes_results}

This task required the robot to locate and grasp small objects placed randomly within the workspace, testing the policy's ability to generalize to novel spatial configurations. Fig.~\ref{fig:pareto_combined}~(b) shows the results.

\textbf{Dominance of Active Vision:}
Consistent with the findings in Task 1, the pure active vision setup~($A$~\textcircled{4}) achieved the fastest execution times (mean \textbf{0.4 min}) and the best success rate (mean \textbf{87.5\%}) with smaller state-vector and computational complexity than multi-sensor baselines. This dominance aligns with recent findings on active perception~\cite{xiong2025visionactionlearningactive}, stating that it can help to provide the IL policy with a focus on the most important object in the scene and thus improve the policy's robustness. Our results indicate that \emph{active perception is more critical for generalization} than static multi-view redundancy.

\textbf{Visual Interference and Hovering:}
Policies using static cameras, either alone or combined with active vision, performed notably worse (e.g.,~$S$,~$S_{\text{L}}WA\text{-}P$~\textcircled{6}). Qualitative analysis revealed a specific failure mode named \textit{hovering behavior}, where the robot freezes or oscillates during the approach phase. Crucially, this behavior was triggered specifically when combining the active stereo-camera with the static wide-angle head cameras. The inclusion of even a single static wide-angle feed alongside the active stream caused the policy to stall and greatly increased the execution time. 
This suggests a \emph{feature conflict} where the ResNet18-based transformer architecture likely struggles to reconcile the dynamic, depth-resolved features of the active view with the conflicting spatial context of the static wide-angle view. Instead of adding information, the co-located static wide-angle camera acts as a distractor, degrading the policy's performance. It remains an open research question whether Vision Transformers (ViT) like DINOv3~\cite{simeoniDINOv32025}, which are known for better feature separation, would exhibit the same sensitivity to this interference.

\section{Synthesis and Practical Design Guidelines}
\label{sec:guidelines}

Combining the insights from both structured and unstructured tasks allows us to derive actionable design guidelines for ACT-based imitation learning on humanoid robots in data-limited regimes:

\begin{enumerate}
    \item \textbf{Active Vision is the Baseline:} A single active stereo-camera ($A$) proved to be the most robust single sensor, offering high generalization capabilities (Task 2) and competitive speed (Task 1). It should be the \emph{default starting point} for humanoid manipulation.
    \item \textbf{Modality Balance is Critical:} Adding tactile sensors ($P$) is not inherently beneficial in small-data regimes. As seen in Task 1, tactile data requires supporting visual context (like wrist cameras) to be interpretable; otherwise, it acts as a noise source due to distribution shifts. Our findings indicate that optimal sensor configurations are highly context-dependent. Consequently, \emph{simply maximizing the number of sensors does not guarantee superior performance}. Moreover, minimizing the sensor set lowers the computational cost. In some cases, benchmarking to identify the most synergistic sensor set could be beneficial.
    \item \textbf{Avoid Co-Located Visual Redundancy:} We observed that combining active and static cameras on the same kinematic link (the head) causes \emph{destructive interference}. Efficient system design should prioritize a single, high-quality active view over redundant static streams that introduce conflicting feature maps.
    \item \textbf{High-Quality Demonstrations are Key:} We observed qualitatively that human demonstrations should incorporate a large enough clearance to obstacles to account for mode averaging.
\end{enumerate}

\section{LIMITATIONS}
\label{sec:limitations}

While this study offers an extensive comparison, several limitations exist:
\begin{itemize}
    \item \textbf{Algorithm Specificity:} These findings are evaluated using ACT. Other leading architectures, such as Diffusion Policy~\cite{chiDiffusionPolicyVisuomotor2024}, handle multi-modal action distributions differently and might exhibit varying robustness to informational noise.
    \item \textbf{Teleoperation Latency:} The VR setup introduced a motion-to-photon latency of 0.5--1.0 seconds. While operators learned to compensate, this may have introduced artifacts (e.g., pauses) in the training data. However, since this latency was constant across the entire master dataset, it cannot explain the variance in performance (e.g., hovering behavior) observed solely between different sensor configurations.
    \item \textbf{Task Diversity:} Both tasks were tabletop manipulation. The optimal sensor set for dynamic tasks, such as locomotion or whole-body lifting, may differ.
    \item \textbf{Data Regime:} We focused on data-limited regimes. The benefit of complex modalities like tactile sensing might only materialize with significantly larger datasets.
    \item \textbf{Sensor Fidelity and Hardware Specificity:} Our findings regarding proprioceptive and tactile modalities are inherently tied to the sensor fidelity of the Unitree G1 (e.g., current-based torque estimation and binary-tending pressure sensors). On high-fidelity robotic platforms with highly accurate joint-torque sensors (e.g., DLR TORO~\cite{englsbergerOverviewTorquecontrolledHumanoid2014}), proprioceptive feedback might prove more beneficial. Our results emphasize that for cost-effective hardware, sensor inclusion must be carefully weighed against the introduced noise.
\end{itemize}

\section{CONCLUSION AND FUTURE WORK}
\label{sec:conclusion}

This extensive benchmark of 14 sensor configurations for humanoid imitation learning \emph{challenges the assumption} that richer sensory data inherently leads to better policy performance. By introducing a \textit{Unified Ablation Framework}, we isolated the impact of sensor selection from human demonstration variance.

Our results hint that in data-limited regimes, the \emph{signal-to-noise ratio of sensory modalities} is a critical determinant of success. We identified that a minimalistic \textbf{active stereo-vision setup} offers a superior trade-off between robustness and complexity, often outperforming extensive multi-sensor arrays. Conversely, additional modalities like tactile pressure were observed to act as distractors when not supported by sufficient visual context or dataset scale.

Ultimately, this study indicates that \emph{hardware design for learning-based robots cannot be decoupled from the data regime}. Future work will investigate the scaling laws of sensor fusion to determine the dataset thresholds required for tactile modalities to transition from noise sources to informative features, potentially leveraging novel synthetic data-generation pipelines like Real2Render2Real~\cite{yuReal2Render2RealScalingRobot2025a} to overcome physical data limitations. Moreover, we hypothesize that the observed trade-offs extend to broader classes of algorithms, including Diffusion Policies (e.g., utilizing 3D representations~\cite{zeGeneralizableHumanoidManipulation2025}) and Vision-Language-Action (VLA) models~\cite{brohanRT2VisionLanguageActionModels2023}. Determining whether the pre-trained representations in VLAs mitigate these noise sensitivities or if they exhibit similar failure modes in data-limited regimes constitutes an interesting potential for future cross-architecture research.

\vspace{1.5cm}

\balance

\bibliographystyle{IEEEtran}
\bibliography{bibtex/masterbib}

\end{document}